\documentclass[10pt, a4paper]{article}

\usepackage[]{lrec-coling2024} 

\usepackage{multirow}
\usepackage{graphicx}
\usepackage{booktabs}
\usepackage{amsmath}

\title{HealthFC: Verifying Health Claims with Evidence-Based \\ Medical Fact-Checking}

\name{Juraj Vladika, Phillip Schneider, Florian Matthes} 

\address{Department of Computer Science, Technical University of Munich \\
         Boltzmannstraße 3, Garching, Germany \\
         \{juraj.vladika, phillip.schneider, matthes\}@tum.de\\}

\abstract{
In the digital age, seeking health advice on the Internet has become a common practice. At the same time, determining the trustworthiness of online medical content is increasingly challenging. Fact-checking has emerged as an approach to assess the veracity of factual claims using evidence from credible knowledge sources. To help advance automated Natural Language Processing (NLP) solutions for this task, in this paper we introduce a novel dataset \textsc{HealthFC}. It consists of 750 health-related claims in German and English, labeled for veracity by medical experts and backed with evidence from systematic reviews and clinical trials. We provide an analysis of the dataset, highlighting its characteristics and challenges. The dataset can be used for NLP tasks related to automated fact-checking, such as evidence retrieval, claim verification, or explanation generation. For testing purposes, we provide baseline systems based on different approaches, examine their performance, and discuss the findings. We show that the dataset is a challenging test bed with a high potential for future use.
 \\ \newline  }

\begin{document}

\maketitleabstract

\section{Introduction}

Health can be defined as "a state of complete physical, mental, and social well-being" and is a popular point of discussion both in everyday life and in online spaces \cite{Kuhn2017-kk}.  
The Internet has made seeking information about personal and public health easier than ever before. Many people have turned to online blogs and news portals as a source of evidence regarding health-related inquiries. According to a report released by the Pew Research Center \cite{fox2013health}, over one-third of American adults have searched online for medical conditions that they might have, and they first consult the Internet before deciding if they should visit a medical professional. 
They mostly try to find answers to their medical questions before going to a medical professional or make decisions about whether to consult a doctor or not.

\begin{table}[t]
\centering
\begin{tabular}{p{\columnwidth}}
\\
\hline
\textbf{Claim:} \small Can regular intake of vitamin C prevent colds?\\
\hline
\textbf{Evidence:} \small \textcolor{violet}{The recommendation to take high-dose vitamin C at the first signs of a cold cannot be confirmed by studies.} If cough, sniffing or sore throat are already present, vitamin C does not seem to have any detectable effect. The daily requirement for the vitamin is about 100 milligrams, with the recommendations slightly fluctuating [2,3]. This amount is contained in an apple, half a pepper or two tomatoes [4]. (...) \\
\hline
\textbf{Verdict:}  \color{red} \textbf{Refuted} \\
\hline
\\
\hline
\textbf{Claim:} Does melatonin help against jet lag? \\
\hline
\textbf{Evidence:} \small This sounds plausible at first because melatonin plays an important role in sleep-wake rhythm [4]. We have found an overview of ten individual studies [1] and a newer individual study [2]. At random, the test subjects received melatonin or a dummy medication. \textcolor{violet}{Overall, the studies show that melatonin may help better against jet lag than a sham drug.}  (...) \\
\hline
\textbf{Verdict:}  \color{teal} \textbf{Supported} \\
\hline
\end{tabular}

\caption{\label{tab:claims}Example of two claims from \textsc{HealthFC} with a snippet of evidence documents and verdicts. Manually annotated evidence sentences are highlighted in \textcolor{violet}{violet}.}
\end{table}

With the increasing volume of new data generated daily and the rapid speed at which information is propagated in digital media, keeping track of trustworthy sources has become challenging. 
This has facilitated the spread of misinformation -- content that is usually false, misleading, or not backed by any relevant knowledge sources. In the period of the COVID-19 pandemic, medical misinformation has led people to turn to unsafe drugs and unproven treatments  \cite{doi:10.1177/0956797620939054, zarocostas2020fight}. The challenge of seeking credible health-related information is further amplified by the advent of digital health assistants and generative language models, which have the ability to generate eloquent responses for any input query yet are prone to "hallucinating" knowledge or omitting important details \citep{10.1145/3571730}. 

The usual way for biomedical researchers to test their hypotheses related to human health is by conducting \textit{clinical trials}. 
Clinical trials are carefully designed research studies that seek to investigate the efficacy and safety of biomedical or behavioral interventions in human subjects, 
which may include novel treatments such as vaccines, drugs, dietary supplements, medical devices, or known interventions that require further examination \citep{piantadosi2017clinical}. When performed with high standards,
clinical trials serve as a high-quality and trustworthy expert-curated source of evidence for health-related decisions. Multiple clinical trials related to the same topic are commonly combined into a \textit{systematic review}. These reviews 
serve as a medical artifact providing guidelines concerning treatments and medical decisions with varying levels of evidence and strength of recommendation \citep{sekhon2017acceptability}.

Fact-checking is the task of assessing factual claims that are contested, using relevant evidence from credible knowledge sources. It is a time-consuming task that is still usually performed manually by dedicated experts in journalism \citep{guo-etal-2022-survey}.
Recently, solutions based on Machine Learning (ML) and Natural Language Processing (NLP) have been developed to automate parts of the fact-checking process. Considering the complexity of the task, current solutions
are still far from achieving human-level performance. Still, they can be used to assist human fact-checkers in their work, such as discovering evidence \citep{Nakov2021AutomatedFF}. 

While multiple datasets for automated fact-checking of health-related and biomedical claims have been constructed in recent years, none of them use clinical studies as their primary source of knowledge to determine a claim's veracity. This is a major drawback, considering the importance of clinical trials and systematic reviews in making health-related decisions in medicine. Furthermore, most datasets provide only top-level labels like "true" and "false" with no information regarding the level of evidence and certainty in the label. Finally, virtually all datasets contain claims solely in English. To address these research gaps, in this paper, we present the following contributions:

\begin{enumerate}
  \item We introduce \textsc{HealthFC}, a constructed bilingual German and English dataset, featuring 750 health-related claims and richly annotated metadata. This includes veracity labels from a team of medical experts, level of evidence, and explanatory documents written in lay language describing clinical studies used for assessment. We additionally provide manually annotated evidence sentences from documents. The dataset enables testing of various NLP tasks related to automated fact-checking.
  \item We develop diverse baseline systems to benchmark the performance of evidence selection and verdict prediction on the dataset and describe the findings and challenges.
  \item We provide additional insight and experiments related to different evidence sources and levels of evidence in the verification process.
\end{enumerate}

We provide the dataset and code in a public GitHub repository.\footnote{\url{https://github.com/jvladika/HealthFC}/}


\section{Related Work}
\subsection{Medical NLP Tasks}
Healthcare is a popular application domain in artificial intelligence and natural language processing. The complexity of language found in sources like biomedical publications and clinical trial reports makes it a challenging domain to work with. To overcome these obstacles, general-purpose NLP models are pre-trained and fine-tuned on domain-specific biomedical and scientific texts. This includes models like SciBERT \citep{beltagy-etal-2019-scibert} and BioBERT \citep{lee2020biobert}.
Biomedical NLP tasks include a wide array of common NLP tasks, such as natural language inference \cite{jullien-etal-2023-nli4ct}, named entity recognition \cite{zhao2019neural}, 
dialogue systems \cite{zeng2020meddialog}, evidence inference \cite{deyoung-etal-2020-evidence}, or text summarization \cite{abacha2021overview}. 

A knowledge-intensive NLP task related to fact-checking is \textit{question answering} (QA). 
In particular, \textit{biomedical question answering} can be divided into four groups: scientific, clinical, examination, and consumer health \cite{jin2022biomedical}. The first three groups target questions that help medical professionals and researchers conduct their work. Examples include PubMedQA \cite{jin-etal-2019-pubmedqa} and BioASQ-QA \cite{bioasq}.
Our work is mostly similar to consumer-health QA, where the goal is to help the general population seek medical advice, and the produced answers should be user-friendly \cite{10.1093/jamia/ocz152}. Recent QA datasets focus on producing long answers for medical inquiries by generative language models, as seen in ExpertQA \cite{malaviya2023expertqa}.
In QA, the task is to answer a specific question, while our dataset more intuitively belongs to automated fact-checking since it assesses claim veracity.

\subsection{Medical Fact-Checking}
Numerous datasets for automated fact-checking have been released in recent years \cite{guo-etal-2022-survey}. 
Most of these datasets are related to society, politics, and general online rumors. Examples include MultiFC \cite{augenstein-etal-2019-multifc} or the Snopes dataset \cite{hanselowski2019richly}, where the authors leveraged existing claims and explanations from professional fact-checking platforms to construct the dataset. We also followed such an approach, focusing on health claims. Most fact-checking datasets contain claims and evidence solely written in English. The only other dataset we found with some claims in German is the multilingual dataset X-Fact \cite{gupta2021x}, which focuses on challenges in cross-lingual transfer for automated fact-checking.

Datasets with biomedical and health-related claims started emerging since 2020 due to online content related to the pandemic of COVID-19 \cite{vladika-matthes-2023-scientific}. These datasets differ with respect to their primary source of claims and evidence. Datasets like SciFact \cite{wadden-etal-2020-fact} and HealthVer \cite{Sarrouti2021Healthver} feature expert-written claims stemming from biomedical research publications and user search queries, respectively, and pair them with abstracts of scientific publications that provide evidence for assessing the claim. On the other hand, datasets COVID-Fact \cite{saakyan-etal-2021-covid} and CoVERT \cite{mohr-whrl-klinger:2022:LREC} take social media posts to gather the claims, the former pairing claims from Reddit with accompanying evidence articles, and the latter taking causative biomedical claims from Twitter posts paired with manually annotated evidence documents from Google search results. Similar to our dataset in terms of construction is PubHealth \cite{kotonya-toni-2020-explainable}, which uses news titles of articles on public health from dedicated fact-checking websites as claims and accompanying article text as the evidence source. Still, the dataset is relatively noisy since the news titles often do not make a factual and atomic claim.
Also related is RedHOT \cite{wadhwa-etal-2023-redhot}, a dataset of medical questions from reddit that matches them with relevant clinical trial reports. 
Recent work by \citet{wührl2024makes} focuses on determining the verifiability of biomedical claims based on claim properties.

To the best of our knowledge, \textsc{HealthFC} is the first dataset for medical fact-checking to use clinical trials and systematic reviews as its main source of evidence.
More precisely, it utilizes knowledge from clinical studies as its initial source and presents it understandably for everyday users in a form of articles. Furthermore, this is the only dataset of health claims to feature the strength of found evidence in its labels and a short explanation paragraph for every verdict decision. It also covers a wider variety of topics concerning all segments of human health, when compared to other datasets focusing only on COVID-19-related claims (as seen in HealthVer, COVID-Fact, and CoVERT).


\section{Dataset Construction}

\subsection{Data Source}
The dataset was constructed from the publicly available data on the web portal \textit{Medizin Transparent}.\footnote{\url{https://medizin-transparent.at/}} It is a project by the team of Cochrane Austria. Cochrane is an international charitable organization formed to organize medical research findings to facilitate evidence-based decisions about health interventions involving health professionals, patients, and policymakers \cite{kleinstauber1996cochrane}.

The team of Medizin Transparent uses a systematic approach to perform fact-checking. The process usually starts with a user inquiry regarding a health-related issue. 
In addition, health claims that are currently trending on popular news portals are considered as well. Then, this inquiry is formed to a precisely defined question that is used to search through several research databases dealing with biomedical research, where relevant studies are manually filtered down.
The preference as a primary source is given to systematic reviews since they present a comprehensive synthesis of results on a research topic in previously published studies. If no systematic reviews are available, the conclusions are drawn from as many informative individual studies as possible to make the best-informed decision. The narrowed-down studies are assessed with regard to quality and significance using previously defined criteria, ensuring the trustworthiness and consistency of the sources. The quality of studies is checked by at least two people from the project's scientific team. The results are summarized by an author, 
checked by a medical professional, and described in a comprehensible and easily understandable way for a wide audience.

 We constructed a scraping project with the Python library Scrapy\footnote{\url{https://scrapy.org/}} and collected all the text from the articles. Because the crawled articles from the portal are exclusively written in German, we translated them into English to provide a wider reach and alignment with similar datasets in English. Claims and explanations of the verdicts were translated with the DeepL API.\footnote{\url{https://www.deepl.com/pro-api}} For the article text, DeepL could not be used due to the character limits of the free API version. Instead, we translated the longer document texts with the Opus-MT library \cite{TiedemannThottingal:EAMT2020}, an open-source tool with a proven record of generating translations of high quality. 
All the translated articles were read by the authors during the evidence annotation process and any spotted mistakes in translation were manually corrected by the authors who are native German and fluent English speakers, to ensure high quality of the provided text.

\subsection{Claims and Labels} 
The two main components of the \textsc{HealthFC} dataset are claims and evidence documents. Each of the 750 claims is paired with a single evidence document. These evidence documents come directly from the fact-checking portal and were written and proofread by the portal's medical team.
The claim veracity labels also come directly from the medical experts.
One specific aspect of the veracity labels in our dataset is that, on top of providing a positive ("true") or negative ("false") label, there is additionally a three-point scale denoting the \textit{level of evidence}. This refers to how strong the findings from clinical trials were and how certain the veracity label of the claim is based on available evidence. The medical team follows internal guidelines on determining which of the three scores of level of evidence to assign to each claim. It is based on the GRADE framework, used for giving clinical practice recommendations \cite{GOPALAKRISHNA2014760}. 

Following other common fact-checking datasets,
we map all the veracity labels to three final high-level verdicts: \textit{supported}, \textit{refuted}, and \textit{not enough information} (NEI). On top of these three labels, there is also a label for the aforementioned level of evidence, in case of the \textit{supported} and \textit{refuted} claims.
In some other datasets, like SciFact, the NEI label signifies that no relevant evidence documents related to the claim are present in the dataset. On the other hand, claims labeled with NEI in our dataset will always be paired with an evidence document. This evidence document usually reports how no relevant clinical studies were found in academic databases or those found are lacking in quality, and therefore, a reliable verdict on the claim's veracity cannot be made. 

\subsection{Evidence Annotation} 
Even though the evidence documents for each claim contain enough information to make a final verdict, not everything in the documents is relevant -- they often contain background information that is interesting for readers, but not necessary to make a final decision. 
Hence, we decided to annotate individual sentences that provide evidence (rationale) for making a final verdict on claim's veracity. 

Two authors
served as annotators. They followed a systematic annotation process by first reading the claim along with its stated verdict and then the full article. All sentences in the article were split automatically using a sentence tokenizer from NLTK. The task was to select only those sentences that make a statement on a claim's veracity. The maximum number of sentences to be selected as rationales was capped at 5. It was empirically determined that rarely are more than 5 sentences needed to make a verdict and this number follows the convention of other fact verification datasets like FEVER \cite{thorne-etal-2018-fact} and COVID-Fact \cite{saakyan-etal-2021-covid}. 
The annotators held regular meetings to discuss and resolve any uncertainties during the labeling process. 

In order to assess the inter-annotator agreement in the labeling process, 50 evidence documents (6.7\%) were selected for mutual annotation by both authors. Cohen's $\kappa$ coefficient \cite{cohen1960} was determined to be 0.72. Cohen suggested interpreting the values of $\kappa$ between 0.61 and 0.80 as substantial agreement. This is comparable to Cohen's $\kappa$ of 0.70 in \citet{hanselowski2019richly} and 0.71 in \citet{wadden-etal-2020-fact}, as well as Fleiss' $\kappa$ coefficient \cite{CTX:C6682} of 0.68 in \citet{thorne-etal-2018-fact} and 0.74 in \citet{hu-etal-2022-chef}.




\section{Dataset Description}
In this chapter, we will provide descriptive analytics and statistics of the dataset, and outline some specific characteristics and challenges. 

\subsection{General Overview of Dataset}
\begin{figure}[h]
  \centering
  \includegraphics[width=\linewidth]{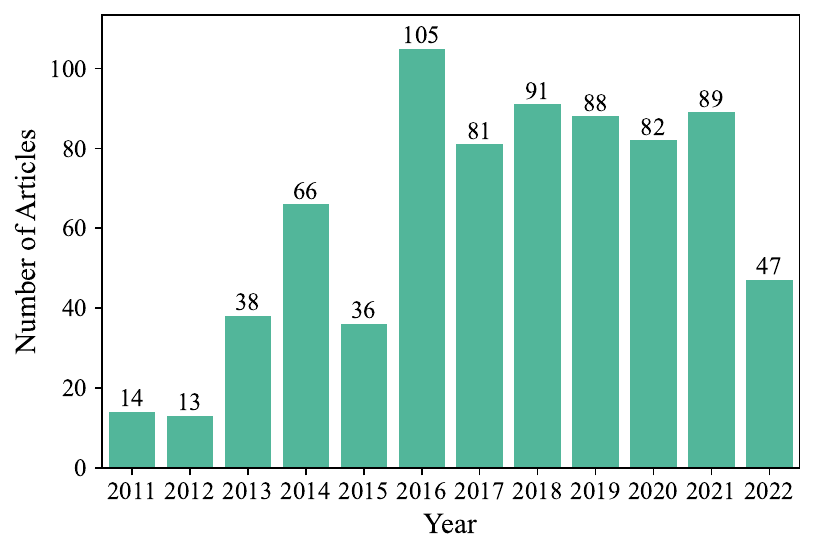}
  \caption{Number of collected health fact-check articles by year of publication.}
  \label{fig:claim-count}
\end{figure}

The \textsc{HealthFC} dataset consists of 750 scientifically fact-checked health claims and evidence articles. Our dataset is available in both English and German. 
Figure~\ref{fig:claim-count} shows the number of yearly published articles over the project's time span. The plotted distribution reveals a significant increase in the number of articles per year, peaking in 2016 with 105 articles. After that, around 80 to 90 claims were fact-checked annually until the number dropped again in 2022. 
Since articles can get outdated with time as new clinical studies are published, the team periodically checks all claims and updates the verdicts as appropriate. Therefore, the knowledge was kept up-to-date with the latest developments by the authors. 
After the release of our dataset, it is possible some verdicts get outdated with time. Therefore, the dataset reflects the latest knowledge up to the end of 2022 and should be used as such.

\begin{figure}[h]
  \centering
  \includegraphics[width=0.99\linewidth]{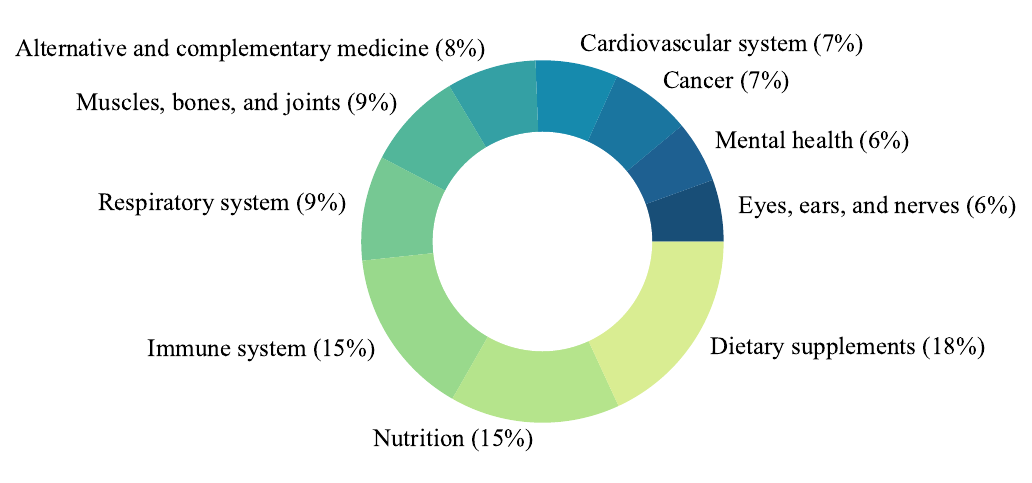}
  \caption{Distribution of the top ten most popular health topics in the collected dataset.}
  \label{fig:domain-donut}
\end{figure}

The dataset covers a diverse range of health topics, encompassing many subdomains. Listing all covered topics would go beyond the scope of this brief dataset description. To gain insight into the most frequently covered subdomains, a subset of the top ten topics is visualized in Figure~\ref{fig:domain-donut}. 
The chart depicts the relative share of fact-checks among the top ten topics.
It is evident that inquiries about eating habits are most popular, since the topics dietary supplements and nutrition account for respectively 18\% and 15\%. Dietary topics are a complex topic for which an abundance of health advice can be found on the Internet.
The third most popular topic is the immune system. It plays a vital role in bodily defense
and is thus responsible for many health conditions. Other prominent subdomains focus on specific body systems, such as the respiratory, musculoskeletal, or cardiovascular systems. Alternative and complementary medicine covers non-traditional forms of healing.

Apart from general health domains that remain consistent over time, the dataset also contains topics that depend on current trends and events. One such topic is COVID-19, which has dominated the news and public health discussions since its outbreak in 2019. The COVID-19 pandemic has impacted people's health and well-being worldwide. It has highlighted the importance of online resources as a primary information medium for people seeking health and medical advice.

\subsection{Descriptive Statistics of Dataset} 
The \textsc{HealthFC} dataset comprises health claims, evidence articles, verdicts, and manually annotated evidence sentences that support the final verdict. Each evidence article also contains an explanation paragraph that serves as a short summary of the article and a justification of the verdict. Table~\ref{tab:summary-statistics} summarizes descriptive statistics of the English texts in our provided dataset. It includes the mean, standard deviation, minimum, and maximum values for various aspects of the dataset. 
An interesting observation is that the word count of explanations has a mean of 40.0
and ranges from only 7 up to 103 words. On the other hand, the word counts of the manually selected evidence sentences are almost twice as high. This demonstrates that, despite limiting the evidence sentences to a maximum of five, the original explanations (summaries) of the verdicts are more concise. These short explanatory summaries could be used for the task of explanation generation in future work.

\begin{table}[h]
  \begin{tabular}{lcccc}
    \toprule
    \textbf{Aspect}&\textbf{$\mu$}&\textbf{$\sigma$}&\textbf{Min}&\textbf{Max}\\
    \midrule
    \small No. of evid. sent. & 3.4 & 1.2 & 1 & 5\\
    \small No. of all sentences & 59.0 & 25.3 & 16 & 168\\
    \small Words: articles & 857 & 369 & 244 & 2677\\
    \small Words: explanations & 40.0 & 18.3 & 7 & 103\\
    \small Words: evid. sent. & 76.6 & 32.2 & 15 & 189\\
  \bottomrule
\end{tabular}
  \caption{  \label{tab:summary-statistics}
 Quantitative statistics of the dataset.}
\end{table}

\begin{figure}[h]
  \centering
  \includegraphics[width=0.99\linewidth]{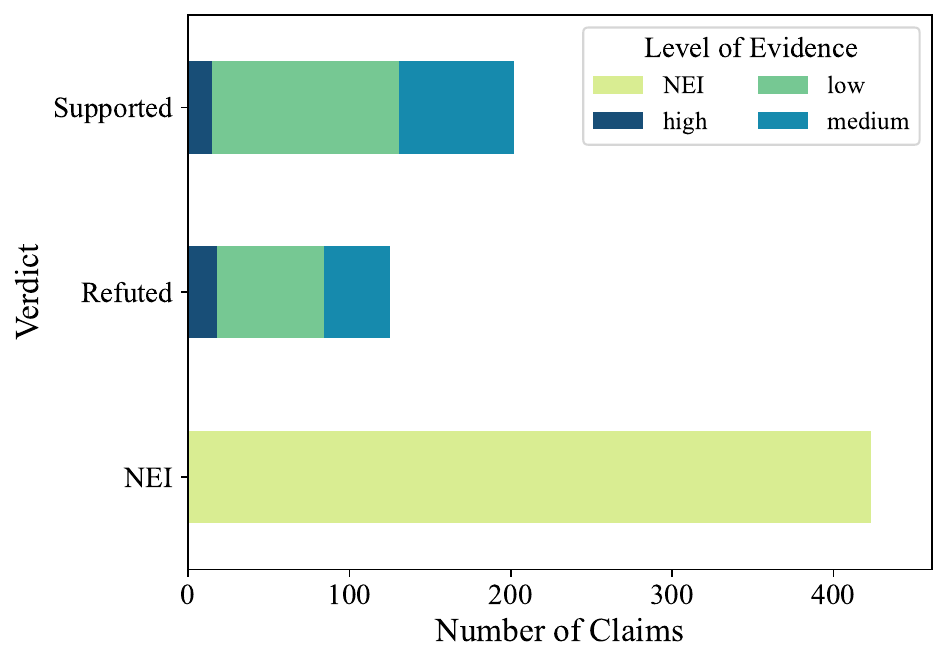}
  \caption{Evidence level count by verdict label. NEI denotes "not enough information".}
  \label{fig:evidence-count}
\end{figure}

The absolute frequency of three verdict labels (refuted, supported, NEI) for all 750 articles is presented in Figure~\ref{fig:evidence-count}. The distribution in the chart is highly skewed towards the NEI verdict, being the majority class with 423 articles. This is to be expected due to the complex nature of research in the health field and due to the strict guidelines followed by Cochrane reviewers for giving a conclusive verdict. Health claims are often subject to ongoing research and clinical trials, and there may not be enough evidence available at the time of the assessment to determine whether a claim is true or false. 

The number of articles where claims are supported (202) is less than half of those in the NEI class, while the fewest belong to the refuted (125) category. Supported and refuted claims additionally have levels of evidence provided on a three-point scale, which refer to the frequency and strength of discovered evidence in clinical studies related to the claim. The distribution of the level of evidence is also skewed towards lower levels. This demonstrates once again the challenge of making decisive health assessments. 
A similar distribution was found by \citet{howick2022most}, who manually analyzed 2,428 Cochrane reviews and found that only 6\% of interventions had high-quality evidence to support the effects, with an additional 20\% having evidence of moderate quality for supported claims.

\begin{table*}[t]
\centering
\resizebox{\textwidth}{!}{
\begin{tabular}{c|c|ccc|ccc|c}
\hline
\multicolumn{2}{l}{} & \multicolumn{3}{c}{\textbf{Evidence Selection}} & \multicolumn{3}{c}{\textbf{Veracity Prediction}} & \textbf{Oracle Ver. Pred.} \\ \hline
\textbf{System} & \textbf{Base Model} & \textbf{Precision} & \textbf{Recall} & \textbf{F1 Macro} & \textbf{Precision} & \textbf{Recall} & \textbf{F1 Macro} & \textbf{F1 Macro} \\ \hline
\multirow{5}{*}{pipeline} & XLM-R (German) & $48.0_{1.5}$ & $48.3_{3.6}$ & $48.1_{2.1}$ & $59.0_{9.2}$ & $59.1_{8.4}$ & $58.4_{9.1}$ & $73.6_{6.5}$ \\
& XLM-R (English) & $51.9_{2.1}$ & $52.9_{4.0}$ & $52.3_{1.4}$ & $64.8_{7.4}$ & $60.0_{5.7}$ & $60.4_{6.1}$ & $74.7_{4.4}$ \\ \cline{2-9}
& BERT & $51.4_{3.4}$ & $51.0_{1.9}$ & $51.2_{1.6}$ & $50.0_{4.8}$ & $50.5_{4.2}$ & $50.1_{4.6}$ & $69.9_{5.6}$ \\
& BioBERT & $52.2_{2.0}$ & $54.4_{2.5}$ & $53.2_{1.0}$ & $62.6_{5.3}$ & $56.3_{4.0}$ & $57.2_{4.2}$ & $78.1_{3.8}$ \\
& DeBERTa & $54.8_{3.7}$ & $56.6_{3.3}$ & $\textbf{55.5}_{1.0}$ & $67.6_{4.6}$ & $64.5_{4.0}$ & $\textbf{65.1}_{3.2}$ & $\textbf{81.9}_{1.4}$ \\ \hline
\multirow{3}{*}{joint} & BERT & $79.1_{3.8}$ & $70.0_{2.1}$ & $73.2_{1.7}$ & $69.4_{4.0}$ & $65.5_{3.5}$ & $66.9_{4.4}$ & --- \\
& BioBERT & $64.2_{2.2}$ & $74.2_{2.8}$ & $67.4_{1.2}$ & $65.1_{3.9}$ & $63.0_{4.2}$ & $63.1_{3.7}$ & --- \\
& DeBERTa & $71.8_{2.8}$ & $75.2_{3.5}$ & $\textbf{73.4}_{1.4}$ & $68.2_{4.6}$ & $66.8_{4.0}$ & $\textbf{67.5}_{3.2}$ & --- \\ \hline
\end{tabular}
}
\caption{\label{tab:results}Results of all baseline systems and models in the form of the mean and standard deviation (subscript) of a 5-fold cross-validation over the dataset.}
\end{table*}

\section{Baselines}
In this section, we introduce the problem and describe baseline systems used to benchmark the performance on evidence selection and veracity prediction on the dataset. We experimented with two types of systems (pipeline and joint) and four different base language models. Pipeline and joint systems are used from \citet{vladika-matthes-2023-sebis}.

\subsection{Problem Statement}
The process of automated fact-checking in our dataset consists of two major components: evidence selection and veracity prediction. 

\noindent \textbf{Evidence selection.} A binary classification task, where given a claim $c$ and an evidence document consisting of $n$ sentences $s_1, s_2, ...,  s_n$, the task is to train a model that predicts $z_i = \mathbf{1}[s_i \: \text{is an evidence sentence}]$. 

\noindent \textbf{Veracity prediction.} A ternary classification task, where for a given claim $c$ and $k$ previously selected evidence sentences $\vec{e} = e_1, e_2, ..., e_k$, the goal is to predict one of the three classes of the final verdicts: $y(c, \vec{e}) \in \{\textsc{Supported}, \textsc{Refuted}, \textsc{Not Enough Info (NEI)}\}$. 

\subsection{Pipeline Systems}
The intuitive approach is to develop two separate models -- one for evidence selection and another for veracity prediction. The evidence sentences selected by the first model are used as input for veracity prediction in the next step. 
It is also common to use the same underlying base model in both steps and fine-tune it for these two different tasks \cite{deyoung-etal-2020-eraser}.

Each candidate sentence $s_i$ from the document is concatenated with the claim $c$ to obtain candidate sequences in the form of $a_i = [s_i; SEP; c]$. Each sequence is encoded with a base language model to obtain their dense representation: $h_i = BERT(a_i)$. This representation is then fed to the classifier model Multi-Layer Perceptron (MLP)
that assigns probabilities on the candidate sentence being evidence: $p_i, \, \bar{p}_i = softmax(MLP(h_i))$. Finally, a selection function labels sentences with a probability over a threshold (fixed at $0.5$) as evidence sentences: $z_i = p_i > 0.5$. In the end, the model selected $k$ final evidence sentences $e_1, e_2, ..., e_k$ as input for the next step. We set $k$ to 5 sentences after experimenting with different values.

The task of veracity prediction is commonly modeled in automated fact-checking as the established task Natural Language Inference (NLI), or more specifically, Recognizing Textual Entailment (RTE), which aims to infer the logical relation (entailment/contradiction/neutral) between a hypothesis and a relation.
In our case, the hypothesis is the claim $c$, and the premise is a concatenation of evidence sentences $e = [e_1; e_2; ...; e_k]$. These two are joined as $x = [c; SEP; e]$ and embedded as $w = BERT(x)$. The final model for sequence classification has to learn the function $\hat{y}(c; e) = softmax(MLP(w))$, which is the probability of a veracity label for the claim $c$ given evidence $e$. The class with the highest probability score is selected as the final verdict $v(c; e) = argmax(y)$.

\subsection{Joint Systems}
Another approach is a system that jointly learns both the tasks of evidence retrieval and veracity prediction. This type of training leverages multi-task learning (MTL) and is beneficial because of data efficiency, reduced overfitting, and faster learning with auxiliary information \cite{Crawshaw2020MultiTaskLW}. 
This is achieved by modeling a unified representation of the claim and the document used for both tasks and a joint loss function that combines the evidence selection loss and veracity prediction loss.

The claim $c$ is concatenated together with all of the sentences $s_1, s_2, ..., s_n$ in the document to obtain a claim+document sequence $seq = [c; SEP; s_1; SEP; s_2; ...; SEP; s_n]$.\footnote{This sequence had to be truncated to $512$ or $1024$ tokens with respect to input limitations, which is good enough for the vast majority of documents in the dataset.} 
This sequence is embedded as $h = BERT(seq) = [h_c; SEP; h_{s_1}; ...; SEP; h_{s_n}]$. The representation of each candidate sentence $h_{s_i} = [h_{w_1}, h_{w_2}, ..., h_{w_m}]$ is singled out from the initial representation and passed to a binary linear classifier that calculates the probabilities of the sentence being evidence: $p_i, \, \bar{p}_i = softmax(MLP(h_{s_i}))$. Those sentences that are above the $0.5$ threshold are selected and used to form the final claim+evidence representation $h_f = [h_c ; h_{e_1}, h_{e_2}, ..., h_{e_k}]$. This representation is given to a ternary classifier that predicts the verdict $v = argmax(softmax(MLP(h_f)))$. 

\subsection{Encoding Models}
To encode the text, we experimented with a number of underlying base models that we found representative of different aspects we wanted to test. BERT \cite{devlin-etal-2019-bert} is used as the representative vanilla pre-trained language model (PLM), which gives a good initial insight into the performance of PLMs on the dataset. BioBERT \cite{lee2020biobert}, an extension of BERT that was fine-tuned to abstracts of biomedical scientific publications, is used to check whether the medical terminology and relations it learned will help assess the claims in this dataset. Additionally, DeBERTa-v3 \cite{he2021deberta}, an improvement of BERT with enhanced training procedure based on disentangled attention, was chosen because it has proven to be powerful for natural language understanding (NLU) tasks, in particular natural language inference and entailment recognition. Finally, XLM-RoBERTa \cite{conneau-etal-2020-unsupervised} is chosen to contrast the performance between the English and German versions of the dataset because it is a powerful multilingual model that was also shown to work well on NLP tasks involving German text \cite{vladika-etal-2022-tum}.

\section{Experiments}
In this section, we report on the experimental setup and results achieved by different systems.

\subsection{Setup}
We performed an array of experiments to test the performance of baseline systems on the two common fact-checking tasks. Considering the dataset's relatively small size, we opted out of declaring a small subset of the dataset to be a test set, but instead split it into 5 folds of equal size and equal label distribution and then performed a 5-fold cross-validation procedure with the final scores being shown with a mean and standard deviation. 
These five splits are released together with the dataset for easier reproducibility. The hyperparameters were mostly the same for all models and setups: learning rate $10^{-5}$, warmup rate $0.06$, weight decay $0.01$, batch size $4$, epochs $7$. For all of the models, their \textit{Large} version was used, imported from the HuggingFace repository. 
The experiments were run on a single Nvidia V100 GPU card with 16 GB of VRAM, for one computation hour per experiment.

\subsection{Results}

The final results of main experiments are shown in Table \ref{tab:results}. The results show the mean and standard deviation over the 5-fold cross-validation of precision, recall, and F1 score, which are useful classification metrics for a dataset with an imbalanced label distribution. All the metrics are macro-averaged scores over the three classes. All experiments were run on the English version of the dataset, except \textit{XLM-R (German)}, which was run on the German version of the dataset. The task of \textit{Evidence Selection} consisted of predicting for each candidate sentence whether it belongs or not to evidence sentences,
with only about 6\% of all candidate sentences had a positive label in this task. The selected sentences in this task are passed over to models in the next task, the \textit{Veracity Prediction}. It is a three-class classification problem with the goal of predicting one of the three classes.
The models used for Veracity Prediction were fine-tuned to predict the label with gold sentences, but during inference time, they used model-selected sentences from the previous step. 
In the last column, \textit{Oracle Verdict Prediction}, we show the scenario where manually annotated ("gold") evidence sentences were used as input to the label-prediction model.

Other than these main experiments, some additional experiments with different settings were performed, in order to test particular challenges in open-domain verification of the dataset. Their results are shown in Table \ref{tab:experiments}. The experiment \textit{Claims only} predicts the veracity by only taking into account the claim text, with no evidence at all. 
For the experiment Google snippets, we ran a search over the Google Search API (on September 1, 2023) with our claims in English and collected the snippets from the first 10 results. These snippets were concatenated as evidence, fine-tuned, and claim veracity  was predicted. This is to test the open-domain claim verification, as Google snippets were used as evidence in other fact-checking datasets \citep{augenstein-etal-2019-multifc, hu-etal-2022-chef}. The experiment \textit{Gold explanations} utilizes explanatory summaries written by authors at the beginning of every fact-check article to check how useful these summaries are for veracity prediction.
All of these experiments used the same 5 folds as the previous table. Finally, the experiment \textit{Level of Evidence} aimed to predict one of the three categories of the level of evidence (\textit{low}, \textit{medium}, or \textit{high}), for which the distribution is shown in Figure \ref{fig:evidence-count}. 

\section{Discussion}
In this section, we discuss the main findings of the experiments, provide deeper insights, perform a qualitative error analysis, and outline the open challenges for future work.

\subsection{Main Results Analysis}
As can be seen in Table \ref{tab:results}, the basic BERT model provides results slightly above 50.0 (F1 macro) for both tasks, which is solid considering the dataset is imbalanced. Still, the biomedical model BioBERT outperforms BERT in both types of systems, which show the benefit of using a domain-specific model for a dataset that includes biomedical terminology. Even though our dataset consists of text written to be understandable to everyday users, it still features a wide array of medical terms and nuances that were probably better captured by BioBERT. Nevertheless, DeBERTa outperformed both BERT and BioBERT in the pipeline system, by a massive margin on veracity prediction. This shows the power of this model for the task of natural language inference in general. This shows that being optimized for good performance in a specific NLP task like entailment recognition can beat a simpler model that is optimized for a specific domain.

When looking at the performance of XLM-RoBERTa for the parallel German and English corpus, it is evident that it worked better for English, especially for evidence sentence selection. This likely stems from the fact that even though the model is multilingual, English was still the most prevalent dataset during pre-training. Still, the results for German are decent while leaving room for improvement. Developing language-specific NLP solutions for a task like this is useful because the speakers of the said language will often seek health advice on the Internet in their native language, so improving the performance on the German version of the dataset remains an open challenge.

The veracity prediction performance with oracle sentences is by far superior to the setup where the model has to select evidence sentences on its own. This shows that detecting appropriate evidence spans and arguments in unstructured text, for a given claim or query, is a challenging problem.
Furthermore, the joint systems show a clear dominance for evidence selection and veracity prediction. Especially for evidence selection, the significant improvement stems from the fact that the learned representation takes the whole document into context and contextualizes sentences to their surroundings. 
The task of veracity prediction is also improved, which shows the clear benefit of multi-task learning and joint task modeling.

\subsection{Qualitative Error Analysis}
To get a deeper insight into the baseline model performance, we do a qualitative error analysis. Table \ref{tab:instance} shows examples where the best-performing DeBERTa model made incorrect predictions. In the first example, even though the baseline model retrieved one of the gold evidence sentences with a refuting conclusion, it also retrieved mentions of earlier studies that supported the claim, which confused the final NLI prediction model. In the second example, only evidence snippets talking about a possible supporting effect were retrieved, which made it hard to predict the gold NEI verdict. In the third example, none of the sentences from the article were above the required threshold to be selected, which led to an incorrect NEI verdict. Improvements to the verification process would include changing the prediction thresholds.

\begin{table}[htpb]
\centering
\small
\begin{tabular}{p{\columnwidth}}
\\
\hline
\textbf{Claim:} \small Are vegetables prepared in a microwave oven less healthy than those prepared in other ways?\\
\hline
\textbf{Gold evidence:}  \scriptsize \textit{If there are health problems related to the microwave, then this is not because the microwave ingredients are destroyed or changed, but because it is simply too unhealthy to eat altogether.}  The week-long feeding with always several times warmed up in the microwave has not led to any signs of poisoning. \\
\textbf{Selected evidence:} \scriptsize Recent studies see the matter more differentiated: Depending on the variety of vegetables and the exact type of preparation, the microwave can sometimes be better and sometimes worse suited than, for example, cooking [2-12]. \textit{If there are health problems related to the microwave, then this is not because the microwave ingredients are destroyed or changed, but because it is simply too unhealthy to eat altogether.} \\
\hline
\textbf{Gold label:}  \color{red} \textbf{Refuted}  \color{black} $\|$  \textbf{Predicted:}  \color{teal} \textbf{Supported}  \\
\hline
\\
\hline
\textbf{Claim:} \small 
Does cat's claw improve joint disease symptoms? \\
\hline
\textbf{Gold evidence:} \scriptsize Whether cat claw helps better in rheumatoid arthritis or osteoarthritis than a placebo \underline{cannot be reliably estimated} on the basis of the available studies. We can't make any statements about effectiveness.\\
\textbf{Selected evidence:} \scriptsize People with joint disease often have a high level of suffering, especially if the medically prescribed products do not help enough. In the case of joint wear (arthrosis) and rheumatoid arthritis, preparations that promise a rapid remedy and are available free of prescription have an economic impact. In laboratory and animal studies there was evidence of a \underline{possible anti-inflammatory effect of cat claw extracts}. As a whole, only a few patients were involved, and very different preparations were tested. This also includes various joint diseases. \\ 
\hline
\textbf{Gold label:}  \color{blue} \textbf{NEI}  \color{black} $\|$  \textbf{Predicted:}  \color{teal} \textbf{Supported}   \\
\hline
\\
\hline
\textbf{Claim:} \small Does brain training boost intelligence? \\
\hline
\textbf{Gold evidence:} \scriptsize Those who train their memory only get better in these exercises, i.e. only in working memory and not in other aspects of intelligence. A review from 2013 casts doubt on how much cognitive training can be helpful for children and adolescents with various mental developmental disorders. \\
\textbf{Selected evidence:} [None] \\
\hline
\textbf{Gold label:}  \color{red} \textbf{Refuted}  \color{black} $\|$  \textbf{Predicted:}  \color{blue} \textbf{NEI} \\
\hline
\end{tabular}

\caption{\label{tab:instance}Examples of claims, with gold and selected evidence snippets, where the DeBERTa baseline made incorrect predictions.}
\end{table}

\subsection{Challenges in Open-Domain Verification}

Table \ref{tab:experiments} shows the results of the additional experiments. For claims only, the classifier is slightly better than random, which shows DeBERTa model utilized its internal world knowledge for predictions but is still considerably worse than any other setup from Table \ref{tab:results}. This shows there are no linguistic patterns spoiling the results, and evidence is needed for a genuine verdict prediction. Fact-checking with Google snippets performs poorly and indicates that these snippets are not informative enough to reach a conclusion, on top of possibly coming from untrustworthy sources. Future work could explore the open-domain claim verification performance using sources such as Wikipedia or PubMed. The performance with explanation summaries is decent but still lacking when compared to using evidence sentences. Future work could utilize these human-written summaries to produce natural language explanations to justify predicted claim verdicts. Finally, the prediction of the level of evidence is also considerably poor and indicates how utilizing this aspect in fact-checking is yet to be explored and refined.

\begin{table}[htbp]
\centering
\begin{tabular}{l|ccc}
\hline
\textbf{Experiment} & \textbf{P} & \textbf{R} & \textbf{F1} \\
\hline
Claims only & $37.4_{2.2}$ & $43.8_{1.0}$ & $39.3_{0.5}$ \\

Google snippets & $44.9_{6.0}$ & $48.0_{2.4}$ & $44.4_{1.6}$ \\

Gold explanations & $46.8_{1.8}$ & $57.4_{2.5}$ & $51.2_{2.1}$ \\
\hline
Level of evidence & $40.3_{0.6}$ & $44.4_{0.8}$ & $42.2_{0.8}$ \\
\hline
\end{tabular}

\caption{\label{tab:experiments}Results of additional experiments with DeBERTa on veracity prediction with different evidence sources and predicting the level of evidence.}
\end{table}

We also refer to the work of \citet{vladika-matthes-2024-comparing} on comparing knowledge sources for open-domain scientific claim verification. In this work, authors test the verification performance on \textsc{HealthFC} and three other datasets using only evidence retrieved from knowledge bases Wikipedia, PubMed, and Google Search. They use only a subset of \textsc{HealthFC} with 327 supported and refuted claims. The results show that evidence retrieved from Wikipedia achieved a binary F1 prediction score 76.5, PubMed 72.0, and Google snippets 74.5. For datasets with complex biomedical claims, PubMed outperformed Wikipedia. This shows that our dataset is better suited for verification over Wikipedia, owing to its focus on common health concerns asked by a wide audience of people. 

As future work, it would be interesting to test the performance of recent Large Language Models (LLMs) on this dataset, especially those specialized for the biomedical domain like PMC-LLaMA \cite{wu2023pmcllama}, Med-PaLM \cite{medpalm}, or BioMistral \cite{labrak2024biomistral}.

\section{Conclusion}
We introduce \textsc{HealthFC}, a novel fact-checking dataset for verifying claims related to everyday health-related concerns. It comprises 750 claims based on users' online inquiries, rich metadata including final verdict labels, explanation paragraphs, full evidence documents, and manually annotated rationale sentences. We describe the dataset creation and collection process in detail and present descriptive statistics.
Finally, we provide results of extensive experiments with two types of baseline systems with multiple base models and show that joint systems with full-document representation outperform the more common pipeline systems. We anticipate that the dataset can help advance the state of automated medical fact-checking and be used for NLP tasks not covered in this paper, such as open-domain verification, explanation generation, and conversational health assistants. 



\section*{Limitations}
Our study employed a rigorous research design that involved the collection, annotation, as well as analysis of a data corpus about fact-checks of health claims. However, it is crucial to acknowledge certain constraints within the study. For one thing, the crawled text data was automatically translated from German to English, which may have resulted in translation errors, especially in view of particular layman's terms or idioms that are difficult to translate. Still, upon manual annotation of evidence sentences, we corrected any spotted errors and inconsistencies. 

Moreover, owing to the German-speaking readership of the Austria-based portal Medizin Transparent, a few articles might focus on topics related to healthcare practices or products popular in Germany, Austria, and Switzerland (e.g., \textit{Does taking LaVita promote health or physical and mental performance?}). Nevertheless, a vast majority of health claims in the dataset talk about common health concerns and can be applied globally without being restricted to a specific country or region. For some health claims, it is possible they were tested with clinical trials featuring a certain demographic subgroup, which could mean they are not applicable on the whole population. This is pointed out in the articles reporting on results. As a last point, it is possible that errors during the dataset annotation have occurred, leading to a not ideal selection of evidence sentences. To mitigate this risk, the annotators discussed their labels in regular meetings to establish a clear understanding of the task.

\section*{Acknowledgements}
This research has been supported by the German Federal Ministry of Education and Research (BMBF) grant 01IS17049 Software Campus 2.0 (TU München). We would like to thank the anonymous reviewers for insightful and helpful feedback.

\nocite{*}
\section*{Bibliographical References}\label{sec:reference}

\bibliographystyle{lrec-coling2024-natbib}
\bibliography{lrec-coling2024-example}

\label{lr:ref}
\bibliographystylelanguageresource{lrec-coling2024-natbib}
\bibliographylanguageresource{languageresource}

\end{document}